# A CHANNEL ATTENTION BASED MLP-MIXER NETWORK FOR MOTOR IMAGERY DECODING WITH EEG

*Yanbin He，Zhiyang Lu，Jun Wang，Jun Shi\**

*School of Communication and Information Engineering, Shanghai University, Shanghai, China.*

## ABSTRACT

Convolutional neural networks (CNNs) and their variants have been successfully applied to the electroencephalogram (EEG) based motor imagery (MI) decoding task. However, these CNN-based algorithms generally have limitations in perceiving global temporal dependencies of EEG signals. Besides, they also ignore the diverse contributions of different EEG channels to the classification task. To address such issues, a novel channel attention based MLP-Mixer network (CAMLP-Net) is proposed for EEG-based MI decoding. Specifically, the MLP-based architecture is applied in this network to capture the temporal and spatial information. The attention mechanism is further embedded into MLP-Mixer to adaptively exploit the importance of different EEG channels. Therefore, the proposed CAMLP-Net can effectively learn more global temporal and spatial information. The experimental results on the newly built MI-2 dataset indicate that our proposed CAMLP-Net achieves superior classification performance over all the compared algorithms.

## 1. INTRODUCTION

Motor imagery (MI) is one of the classic paradigms in the brain-computer interface (BCI) to decode users' intentions through electroencephalography (EEG) [1, 2]. Various machine learning methods, including deep learning (DL) approaches, have been proposed for EEG-based MI decoding [3, 4].

The DL-based methods can be divided into two categories: one applies the deep neural networks, such as stacked auto-encoder and deep belief networks, to the hand-crafted features to further improve the feature representation [5, 6], and the other establishes DL models, such as convolutional neural networks (CNNs), in an end-to-end manner to deal with the raw EEG signals [7]. The latter has attracted considerable attention in recent years, due to the unique advantage of eliminating the need for handcrafted feature extraction.

These CNN-based methods for MI decoding usually compact the EEG signals along spatial channels into vectors and effectively capture the temporal features by convolution operation [8, 9]. However, they generally suffer from the limitation in perceiving global dependencies from both the temporal and spatial views because of the local nature of convolution [10]. Consequently, some crucial information related to long-range dependencies are ignored in feature learning. Therefore, it still has great room to improve the DL-based methods for MI decoding.

More recently, the classical multi-layer perceptron (MLP) has come back as a robust alternative to CNN [11, 12]. The newly proposed MLP-Mixer model, which consists of several stacked Mixer layers, attains competitive performance on image classification benchmarks [11]. Different from the convolution operation in CNN, the novel Mixer layers in MLP-Mixer only rely on the repeated implementation of Layer Norm and MLP either in spatial locations or feature channels. This conceptually and technically simple architecture allows the global spatial features communication between different special locations, and then the long-range dependencies of entire images are covered for the following processing. The MLP-Mixer thus has great potential to explore the advantage of the new MLP-based networks to capture the global dependencies in EEG signals for MI decoding.

On the other hand, different brain regions have different degrees of correlation with MI tasks [13]. However, the current CNN-based works mainly give equal treatment to all EEG channels and inevitably neglect the importance of some specific channels for MI tasks. Consequently, it cannot make full use of the spatial information of EEG. It is worth noting that the MLP-Mixer only consists of several cascaded fully connected components, and thus it cannot discriminate the importance of different EEG channels.

The attention mechanism is an effective strategy for feature selection, which guides the network to focus on the salient parts [14]. Since the different EEG channels contribute differently in MI decoding, the attention mechanism can automatically pick out the important channels and improve their contributions for classification. Hence, the attention mechanism should skip redundant information and benefit the learning of useful spatial features in MLP-Mixer for the EEG-based MI decoding task.

In this work, we propose a channel attention based MLP-Mixer network (CAMLP-Net) for MI decoding with EEG. The novel CAMLP-Net is mainly stacked by several independent CAMLP blocks, each of which is composed of a channel attention unit (CAU) and a time mixing unit (TMU). The former aims to learn the fine-grained spatial information, while the latter captures global temporal information. Therefore, CAMLP-Net can effectively learn both global temporal and spatial information for MI decoding. The experimental results on a public dataset indicate the superior performance of CAMLP-Net.

The main contributions of this work can be summarized as follows:
1) We propose to apply the newly developed MLP-based architecture to the EEG-based classification task. Specifically, in CAMLP-Net, each EEG channel is treated as the individual patch in the original MLP-Mixer for further linear projection, which can efficiently capture the global temporal information to promote feature representation. To the best of our knowledge, this is the first work in the field of signal classification based on the MLP-Mixer architecture.
2) We further develop a channel attention based MLP-Mixer by embedding the attention mechanism into MLP-Mixer. This adaptive channel selection strategy can effectively exploit the importance of different EEG channels, and skip redundant information to promote spatial feature representation. Therefore, the proposed CAMLP-Net can learn more global temporal and spatial information.

## 2. METHODOLOGY

Figure 1 illustrates the architecture of the proposed CAMLP-Net, which contains three modules: 1D-CNN-based local encoder module, CAMLP-Mixer module, and classifier module. The 1D-CNN-based local encoder is applied to obtain the local temporal information for each channel of the raw EEG slices. In the CAMLP-Mixer module, CAU is employed to gain the global special information and exploit the importance of different EEG channels, while TMU is applied to learn the long-range temporal information. The classifier module including global average pooling and linear layer is used to predict decoding category for slice samples.

The general pipeline of the proposed CAMLP-Net for the EEG-based MI decoding task is described as follows:
1) During the training stage, all trial samples in the training set are divided into slices for training CAMLP-Net model. The shuffle strategy is then applied to these slices before feeding them into the CAMLP-Net. The trained CAMLP-Net model thus performs MI decoding task on each multi-channel EEG slice.
2) During the inference stage, each trial sample is segmented into multiple slice data and fed together into the trained CAMLP-Net model to obtain the prediction of each slice. The outputs from CAMLP-Net model are integrated by ensemble scheme for MI decoding result of the trial sample.

The core parts of our proposed approach, including the local encoder and the CAMLP-Mixer module, will be presented in detail in the following sections.

### 2.1. Local encoder

Considering the randomness, poor spatial resolution, and signal-to-noise ratio of EEG signals, it is challenging to directly feed them into the MLP-based network for classification. Thus, the local encoder is applied for preliminary embedding to gain the latent representation.

The 1D-CNN-based local encoder consists of three convolution layers, and each convolution layer is followed by a batch normalization layer. The number of convolution kernels with kernel size of $k$ for the three layers are [$n$, $2n$, $4n$]. Given an input representation $x \in \mathbb{R}^{C \times T}$, where $C$ represents the number of EEG channels, and $T$ represents the number of input sample points. The average pooling with a kernel size of $k$ is applied between the second and the last convolution layer to function the time sequence for each channel of $x$ and maps: $\mathbb{R}^T \to \mathbb{R}^L$, $L = \frac{T}{k}$, where $L$ indicates the size of the time dimension after the local encoder. It should be noted that the local feature encoding only aims to enrich the local representations of EEG in the time domain. Therefore, the number of channels is always kept constant throughout this process to ensure that the spatial structure information is not scrambled.

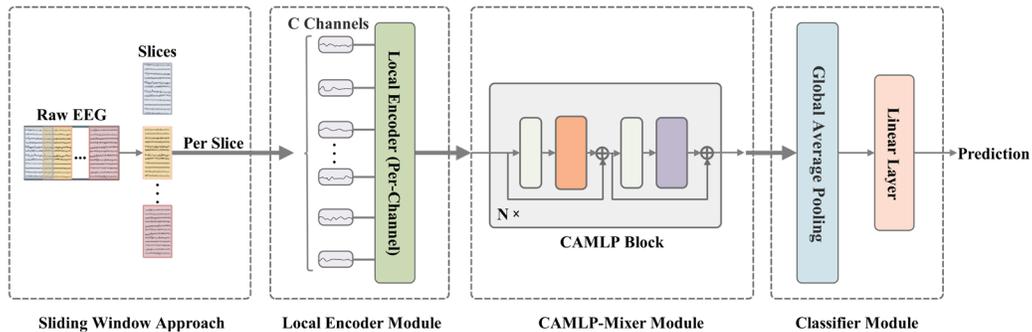

**Fig. 1.** The overall architecture of our proposed CAMLP-Net. *C* represents the number of EEG channels, and *N* indicates the number of CAMLP blocks.

## 2.2. Channel attention MLP-Mixer module

After the preliminary embedding of the local encoder above, the local features of each channel are obtained, but the global temporal information and channel correlation are still lacking. To capture the long-range dependencies and optimize the spatial information of EEG signals, the CAMLP-Mixer module applies MLP-based architecture to capture global spatial representations and temporal information, respectively. Moreover, to skip redundant information of EEG channels, an attention-based mechanism is adopted for channel features selection.

The overview of the CAMLP-Mixer module proposed in this paper is shown in Fig.2. It is stacked by $N$ CAMLP blocks, each of which is composed of two Layer Norm layers, a CAU, and a TMU. For convenience, the basic Mixing Unit (MU), including two fully connected layers and a nonlinear activation function, can be expressed as:

$$MU(x) = W_1(\sigma(W_2 x + b_2)) + b_1 \qquad (1)$$

where $\sigma(\cdot)$ represents the LeakyReLU activation function, $W_1$, $W_2$ indicates the weights of the linear projections and $b_1$, $b_2$ are the biases.

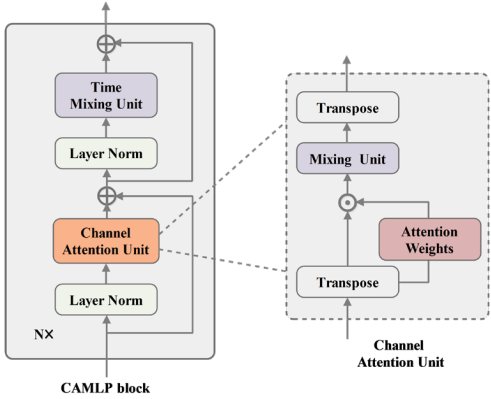

**Fig.2.** Overview of the CAMLP-Mixer module.

The CAU allows temporal features to communicate between different EEG channels. To obtain the information of channel dimension, the transpose of the input matrix is necessary. Concretely, given a feature map $x \in \mathbb{R}^{C \times L}$ generated by the local encoder, $x$ was transposed to $x \in \mathbb{R}^{L \times C}$ to facilitate the process of mixing projection in the channel dimension. To gain the connections between different channels, the mixing projection means the number of channels of $x$ maps: $\mathbb{R}^C \to \mathbb{R}^D \to \mathbb{R}^C$, where $D$ represents the extended dimension. The CAU can be formulated as:

$$z_1 = t(LayerNorm(x^T)) \qquad (2)$$

$$y_1 = x + MU(z_1)^T \qquad (3)$$

$t(\cdot)$, which stands for dot product, is a key ingredient in the above formulation. It is an operation set up to exploit the importance of different EEG channels. Specifically, a set of weight parameters $t \in \mathbb{R}^{1 \times C}$ initialized by Kaiming initialization [15] is assigned for each EEG channel. These adaptive channel weights are learned to enhance the presence of key channels in MI and suppress channels that have low relevance to the identification results. When $t_i = t_j$, $i, j \in \{1,...,C\}$, the expression will degenerate into the basic MU, which means that all EEG channels contribute the same to the classification result.

The TMU allows information to communicate between the local features of all different moments in the same channel. As is well-known that the imaginative behavior of the brain is a complete process, the temporal parts of the EEG are correlated. Therefore, TMU is applied to make full use of the relationship between any two parts of the whole time series. The process of TMU can be expressed as:

$$z_2 = LayerNorm(y_1) \qquad (4)$$

$$y = y_1 + MU(z_2) \qquad (5)$$

where $y_1$ is the output of the CAU, and the output $y$ for subsequent processing has the same shape as the input $x \in \mathbb{R}^{C \times L}$. Moreover, in TMU, the time feature dimension is converted: $\mathbb{R}^L \to \mathbb{R}^H \to \mathbb{R}^L$, where $H$ indicates the hidden dimension.

## 3. EXPERIMENTS AND RESULTS

### 3.1. Dataset and pre-processing

The CAMLP-Net was evaluated on the newly built MI-2 dataset, which consists of 25 right-handed healthy subjects [16]. In this dataset, there are three different tasks of imaging different joint movements of the same limb (imagining grasp movement, imagining right elbow movement, and keeping resting state with eyes open). The EEG signals of these tasks are recoded using the 64-channel gel electrode cap of the standard 10/20 system, and the validity period of each record is 4s. In this study, we adopt the pre-processed dataset. These pre-processing include a band-pass filter and a 50 Hz notch filter. In addition, taking account of the computational cost, these data are also down-sampled to 200 Hz, thus the number of sample points of the original EEG signal is 800 of each channel.

Based on the data usage advice provided by the reference [17], the sliding window approach was applied to segment the data of 7 sessions for each subject (5 sessions of which contains 20 trials for every movement imagination and 2 sessions each include 50 trials of resting state). The size of the sliding window adopted in this experiment was 150 points (0.75s), and there was an overlap of 10 points in adjacent fragments. After that, we utilized the z-score standardization to relieve the fluctuation and non-stationarity. After the above processing, the shape of each slice was $\mathbb{R}^{C \times T}$, where $C$ equals 62 representing the number of electrodes (except HEO, VEO), and $T$ indicates the number of sample points.

### 3.2. Experimental setting

To evaluate the performance of the proposed CAMLP-Net, we compared it with the following approaches:
1) EEGNet [18]: It is a classical compact fully convolutional network with depth-wise and separable convolutions for EEG classification using CNN.
2) TSCeption [19]: A network applies two types of convolutional learners to obtain representation in both time and channel dimensions by its multi-scale convolutional neural network.
3) DeepConvNet [20]: The DeepConvNet employs four convolution-pooling blocks that give it a strong ability to extract features from raw EEG signals.
4) Corr+CNN [17]: This network calculates the correlation matrixes composed of correlation coefficients among all electrodes, and then uses two convolution layers to learn the overall representation of different channels. It is by far the most advanced model on the MI-2 dataset.

Besides, to verify the effectiveness of the channel attention mechanism, an ablation study was also performed to compare the CAMLP-Net and the MLP-Mixer combined with the local encoder.

The 5-fold cross-validation of category balance was applied to all algorithms in our experiment. We ensured that all slices of same trails were divided into either training sets or validation sets when applying the sliding window approach. For fair comparison, these algorithms were performed on the same sliding window approach, and the category of each trail sample was also obtained by ensemble strategy.

In our implementations, the SGD algorithm with momentum of 0.9 was applied to optimize cross-entropy loss function. The learning rate was set to 0.001, and the batch-size was 64. Besides, in the local encoder, there were 3 convolution layers, the kernel size and pooling size $k$ were both set to 3, and the number of the filter $n$ was 4. In the CAMLP-Mixer module, there were 4 CAMLP blocks, the channel extended dimension $D$ was set to 256, while the hidden dimension $H$ in the time domain equaled 128.

### 3.3. Experimental results

Table 1 summarizes the experimental results of different algorithms. It can be found that MLP-Mixer and CAMLP-Net outperform all the compared algorithms, which indicates that it is effective to apply MLP-based architecture to EEG classification. The CAMLP-Net achieves the best mean accuracy of 77.44 ± 0.57% and the F1 score of 77.40 ± 0.58%, which gets the improvement by at least 2.41% on the classification accuracy over other compared algorithms.

Moreover, it is worth mentioning that CAMLP-Net improves by 0.93% and 0.91% on the accuracy and F1 score compared with MLP-Mixer, because the CAMLP-Net applied the channel attention mechanism to obtain more useful spatial information. Therefore, it can be concluded that the CAMLP-Net can effectively classify MI tasks through global information communication and channel attention mechanism.

**Table 1**.
Comparison with representative methods on MI-2 dataset.

| Methods | Accuracy (%) | F1 score (%) |
|---|---|---|
| EEGNet | 70.44 ± 1.18 | 69.90 ± 1.41 |
| TSception | 71.97 ± 0.82 | 71.88 ± 0.88 |
| DeepConvNet | 73.07 ± 1.44 | 73.06 ± 1.27 |
| Corr+CNN | 75.03 ± 3.37 | N/A |
| MLP-Mixer | 76.51 ± 0.77 | 76.49 ± 0.77 |
| CAMLP-Net | **77.44 ± 0.57** | **77.40 ± 0.58** |

We also studied the parameter sensitivity of the CAMLP blocks. Fig.3. shows the performance under different CAMLP blocks, which indicates that our algorithm is generally stable. In addition, our algorithm performs best when the number of blocks is 4, and even the model with the least blocks performs better than other compared algorithms.

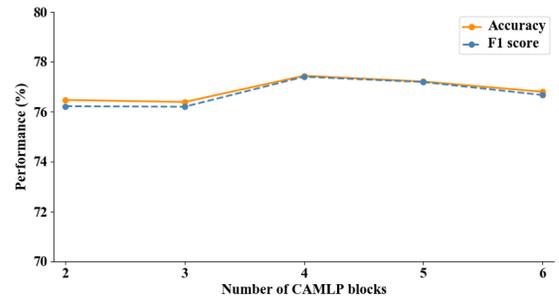

**Fig.3.** The performance of different CAMLP blocks.

### 4. CONCLUSION

In this paper, we propose a channel attention based MLP-Mixer network (CAMLP-Net) for MI decoding, in which the newly developed MLP-based architecture is applied to the EEG-based classification tasks. Besides, we develop a channel attention based MLP-Mixer for global information communication and channel features selection of EEG. The experimental results indicate the effectiveness of our proposed methods.

In the future, we will further explore the application of novel methods in computer vision to the field of EEG classification domain.

# 5. REFERENCES

[1] R. Abiri, S. Borhani, E. W. Sellers, Y. Jiang, and X. Zhao, "A comprehensive review of EEG-based brain–computer interface paradigms," *Journal of neural engineering*, vol. 16, no. 1, 2019.

[2] B. He, H. Yuan, J. Meng, and S. Gao, "Brain–computer interfaces," *Neural engineering*, pp.131-183, 2020.

[3] F. Lotte, M. Congedo, A. Lécuyer, F. Lamarche, and B. Arnaldi, "A review of classification algorithms for EEG-based brain–computer interfaces," *Journal of neural engineering*, vol. 4, no. 2, 2007.

[4] A. Al-Saegh, S. A. Dawwd, and J. M. Abdul-Jabbar, "Deep learning for motor imagery EEG-based classification: A review," *Biomedical Signal Processing and Control*, vol. 63, 2021.

[5] X. Tang, T. Wang, Y. Du, and Y. Dai, "Motor imagery EEG recognition with KNN-based smooth auto-encoder," *Artificial intelligence in medicine*, vol. 101, 2019.

[6] Y. Chu, X. Zhao, Y. Zou, W. Xu, J. Han, and Y. Zhao, "A decoding scheme for incomplete motor imagery EEG with deep belief network," *Frontiers in neuroscience*, vol. 12, 2018.

[7] H. Altaheri, G. Muhammad, M. Alsulaiman, S. U. Amin, G. A. Altuwaijri, W. Abdul, M. A. Bencherif, and M. Faisal, "Deep learning techniques for classification of electroencephalogram (EEG) motor imagery (MI) signals: a review," *Neural Computing and Applications*, pp. 1-42, 2012.

[8] S. Sakhavi, C. Guan and S. Yan, "Learning Temporal Information for Brain-Computer Interface Using Convolutional Neural Networks," *IEEE Transactions on Neural Networks and Learning Systems*, vol. 29, no. 11, pp. 5619-5629, 2018.

[9] S. U. Amin, M. Alsulaiman, G. Muhammad, M. A. Mekhtiche, and M. S. Hossain, "Deep Learning for EEG motor imagery classification based on multi-layer CNNs feature fusion," *Future Generation computer systems*, vol. 101, pp. 542-554, 2019.

[10] Y. Song, X. Jia, L. Yang, and L. Xie, "Transformer-based Spatial-Temporal Feature Learning for EEG Decoding," *arXiv preprint arXiv*: 2106.11170, 2021.

[11] I. Tolstikhin, N. Houlsby, A. Kolesnikov, L. Beyer, X. Zhai, T. Unterthiner, J. Yung, A. Steiner, D. Keysers, J. Uszkoreit, M. Lucic, and A. Dosovitskiy, "Mlp-mixer: An all-mlp architecture for vision," *arXiv preprint arXiv*: 2105.01601, 2021.

[12] X. Ding, X. Zhang, J. Han, and G. Ding, "RepMLP: Re-parameterizing Convolutions into Fully-connected Layers for Image Recognition," *arXiv preprint arXiv*: 2105.01883, 2021.

[13] S. Liu, X. Wang, L. Zhao, B. Li, W. Hu, J. Yu, and Y. Zhang, "3DCANN: A Spatio-Temporal Convolution Attention Neural Network for EEG Emotion Recognition," *IEEE Journal of Biomedical and Health Informatics*, 2021.

[14] W. Tao, C. Li, R. Song, J. Cheng, Y. Liu, F. Wan, and X. Chen, "EEG-based emotion recognition via channel-wise attention and self attention," *IEEE Transactions on Affective Computing*, 2020.

[15] K. He, X. Zhang, S. Ren, and J. Sun, "Delving deep into rectifiers: Surpassing human-level performance on imagenet classification," *IEEE International Conference on Computer Vision*, pp. 1026-1034, 2015.

[16] X. Ma, S. Qiu, and H. He, "Multi-channel EEG recording during motor imagery of different joints from the same limb," *Scientific data*, vol. 7, no. 1, pp. 1-9, 2020.

[17] X. Ma, S. Qiu, W. Wei, S. Wang, and H. He, "Deep channel-correlation network for motor imagery decoding from the same limb," *IEEE Transactions on Neural Systems and Rehabilitation Engineering,* vol. 28, no. 1, pp. 297-306, 2019.

[18] V. J. Lawhern, A. J. Solon, N. R. Waytowich, S. M. Gordon, C. P. Hung, and B. J. Lance1, "EEGNet: a compact convolutional neural network for EEG-based brain–computer interfaces," *Journal of neural engineering*, vol. 15, no. 5, 2018.

[19] Y. Ding, N. Robinson, Q. Zeng, D. Chen, A. A. P. Wai, T. S. Lee, and C. Guan, "Tsception: a deep learning framework for emotion detection using EEG," *IEEE International Joint Conference on Neural Networks (IJCNN).* pp. 1-7, 2020.

[20] R. T. Schirrmeister, J. T. Springenberg, L. D. J. Fiederer, M. Glasstetter, K. Eggensperger, M. Tangermann, F. Hutter, W. Burgard, and T. Ball, "Deep learning with convolutional neural networks for EEG decoding and visualization," *Human brain mapping*, vol. 38, no. 11, pp. 5391-5420, 2017.